\documentclass[10pt,twocolumn,letterpaper]{article}

\usepackage{wacv}              
\usepackage[accsupp]{axessibility} 

\definecolor{wacvblue}{rgb}{0.21,0.49,0.74}
\usepackage[pagebackref,breaklinks,colorlinks,allcolors=wacvblue]{hyperref}

\usepackage{bbding}
\usepackage{multirow}
\usepackage{graphicx}

\def\wacvPaperID{705} 
\def\confName{WACV}
\def\confYear{2026}

\title{Multi-Modal Soccer Scene Analysis with Masked Pre-Training}

\author{Marc Peral$^{1,2}$ \quad Guillem Capellera$^{1,2}$ \quad Luis Ferraz$^{2}$\quad Antonio Rubio$^{2}$\quad  Antonio Agudo$^{1}$  \\  \textcolor{gray}{$^1$Institut de Robòtica i Informàtica Industrial, CSIC-UPC \quad  $^2$Kognia Sports Intelligence}}

\begin{document}
\maketitle
\begin{abstract}
In this work we propose a multi-modal architecture for analyzing soccer scenes from tactical camera footage, with a focus on three core tasks: ball trajectory inference, ball state classification, and ball possessor identification. To this end, our solution integrates three distinct input modalities (player trajectories, player types and image crops of individual players) into a unified framework that processes spatial and temporal dynamics using a cascade of sociotemporal transformer blocks. Unlike prior methods, which rely heavily on accurate ball tracking or handcrafted heuristics, our approach infers the ball trajectory without direct access to its past or future positions, and robustly identifies the ball state and ball possessor under noisy or occluded conditions from real top league matches. We also introduce CropDrop, a modality-specific masking pre-training strategy that prevents over-reliance on image features and encourages the model to rely on cross-modal patterns during pre-training. We show the effectiveness of our approach on a large-scale dataset providing substantial improvements over state-of-the-art baselines in all tasks. Our results highlight the benefits of combining structured and visual cues in a transformer-based architecture, and the importance of realistic masking strategies in multi-modal learning.
\end{abstract}

\section{Introduction}
\label{sec:intro}

Understanding the flow of a soccer match from raw video is a longstanding challenge in sports analytics. Several recent methods attempt to improve tactical coaching and player decision-making through artificial intelligence tools~\cite{wang2024tacticai}. Central to this challenge is the ability to reason about the ball, either its trajectory~\cite{komorowski2019deepball, sorano2020automatic, amirli2022prediction, kim2023ball, capellera2024footbots, capellera2024transportmer, li2025spatiotemporal, soccer3D, capellera2025unified}, its possessor~\cite{link2017individual, kim2023ball, yamamoto2024theory, peral2025temporally} or which action is the possessor performing with it~\cite{sanford2020group, sorano2020automatic, capellera2024transportmer, peral2025temporally, ochin2025game, majeed2025real, rao2024matchtime, rao2025multi, rao2025towards, giancola2023towards}. While player tracking systems are increasingly available and reliable~\cite{cioppa2021camera, somers2024soccernet, cioppa2022scaling, cui2023sportsmot,perezyusWACV22}, direct ball tracking remains error-prone, especially in monocular footage~\cite{komorowski2019deepball, majeed2025real}. Occlusions, limited resolution, and unpredictable motion make it difficult to acquire accurate ball annotations at scale, motivating the development of models that can infer ball dynamics from player behavior alone~\cite{amirli2022prediction, kim2023ball, capellera2024transportmer, xu2024sports}.

Recent works have addressed this task either by using handcrafted rules based on proximity and speed~\cite{link2017individual, vidal2022automatic}, or by modeling player-ball interactions with structured learning frameworks~\cite{kim2023ball, capellera2024transportmer, capellera2025unified, majeed2025real}. However, most existing methods rely on strong assumptions like access to precise positional or velocity information, or rigid task pipelines that require supervision in other partial tasks and add sources of errors. These limitations reduce robustness in real-world settings, where tracking noise, visual occlusions, or poor sensor coverage are common~\cite{gutierrez4998149pnlcalib}.

In this work, we propose a multi-modal and multitask model that jointly 1) infers the trajectory of the ball, 2) classifies the ball state at each frame, and 3) identifies the player in possession of the ball, using only player-centric information. Our method fuses structured inputs like trajectories and player types with visual features extracted from image crops around players to capture their spatial context. We incorporate these heterogeneous inputs into a unified transformer-based architecture, which models both temporal evolution and social interactions across agents. By treating all prediction tasks as supervised and parallel, our approach avoids rigid pipelines and allows the model to learn shared representations that benefit all tasks jointly.

In addition, we introduce CropDrop, a tailored masking-based pre-training strategy inspired by successes in Natural Language Processing (NLP)~\cite{devlin2019bert, wettig2022should, yang2022learning, liang2024bpdec, zheng2025exlm} and vision representation learning~\cite{dosovitskiy2020image, bao2021beit, he2022masked, tong2022videomae, bachmann2022multimae}. Unlike standard random masking, our approach targets entire visual sequences to simulate realistic occlusion or dropout scenarios. This forces the model to rely on structured context to infer missing visual cues, improving generalization. We show that this strategy is particularly effective in our multi-modal setting, where the alignment between visual and structured inputs is essential for high-quality representation learning.

Crucially, our setup assumes no access to ball information during inference, and only uses player-related data as input. This constraint makes the inference problem significantly more challenging and realistic, particularly for non-instrumented or broadcast settings. Moreover, we demonstrate that combining visual and structured data leads to significantly more robust and accurate predictions than either modality alone. Then, our contributions are listed as:
\begin{itemize}
    \item We present a unified multi-task model that jointly predicts ball trajectory, ball state, and ball possessor, all of them just from player-centric inputs.
    \item We demonstrate that incorporating visual information, enabled through a novel masking-based pre-training strategy called CropDrop, significantly helps the model's converging towards more optimal predictions.
    \item We evaluate our method on a large-scale dataset of professional soccer matches, proving stronger performance than competing approaches on all tasks.
\end{itemize}

\section{Related Work}
\label{sec:relatedwork}
In this section, we review prior work related to the tasks our model addresses: 1) identifying the ball possessor, 2) inferring the ball trajectory and 3) classifying the ball state. Finally, we introduce previous work that applied a masking.

\noindent\textbf{Possessor Identification.} Several works have explored pass receiver prediction in soccer~\cite{spearman2017physics, xie2020passvizor, dick2022can, honda2022pass, rahimian2023pass, wang2024tacticai}, often framing it as a key component of broader tactical understanding. Some of these works suggest ball possessor identification as a potential downstream application of their models. However, a key limitation of these methods is that they assume knowledge of the exact moment when a pass is received, making them unsuitable for continuous possessor tracking. Furthermore, many models focus exclusively on completed passes, either successful or not, thus excluding cases of incomplete or interrupted plays where there is no clear ball possessor, such as deflections, bounces, or duels. As a result, these approaches are not directly applicable to frame-level possessor identification in real-world conditions as we do in this paper.

Alternative approaches that directly try to handle possessor identification have relied on combining player and ball trajectories with handcrafted distance-based heuristics to determine the ball possessor at each frame~\cite{link2017individual, vidal2022automatic}. These methods are highly dependent on having smooth high-resolution positional data, particularly for the ball, which is often compromised to guarantee in real-world tracking scenarios. Kim \etal~\cite{kim2023ball} move away from the need for explicit ball tracking by predicting the ball trajectory just from player motion. While this approach is promising, their method assumes the availability of precise velocity signals for each player, which are difficult to estimate reliably when tracking data are noisy or sparse. \cite{peral2025temporally} focuses on identifying ball-related events by first detecting the ball possessor. Their approach uses visual input in the form of player-centric video tubes, that is, cropped image sequences following player bounding boxes over time. While this method shows robustness to missing or noisy tracking data, it suffers when the ball is occluded or outside the camera frame, particularly in crowded scenes.

Our proposed method addresses these limitations by fusing tracking and visual information. By jointly adopting structured trajectory features with image-based cues, our model can reliably identify the ball possessor without requiring highly accurate position or velocity data, and remains robust even when the ball is partially or fully occluded in the visual modality.

\begin{figure*}[t!]
  \centering
   \includegraphics[width=1\linewidth]{figures/overview.pdf}
   \vspace{-0.6cm}
   \caption{\textbf{Overview of our model to jointly infer ball possessor, ball state and ball trajectory from player trajectories, player type and image crops.}  After the three input modalities are projected to a common embedding, concatenated (c) and fused, two CLS tokens are appended to aggregate global information specific for their supervised tasks. The coarse and fine encoders are formed by Set Attention Blocks that act both socially and temporally. Following the splitting (s) of the tokens, different MLPs guide them to become the proper outcomes to the three tasks. A switch represents that the agent masking strategy CropDrop is only used during pre-training.}
   \label{fig:overview}
\end{figure*}

\noindent\textbf{Trajectory Inference.} Trajectory forecasting~\cite{zheng2016generating, felsen2018will, zhan2018generating, yeh2019diverse, capellera2024footbots, capellera2025unified} and imputation~\cite{cao2018brits, liu2019naomi, qi2020imitative, omidshafiei2021time, xu2023uncovering, everett2023inferring, choi2024trajectory, xu2024sports} assume partial knowledge of the target agent's history, but that reliance on past or future information compromises those models performance for long sequences~\cite{gu2017non, lee2018deterministic}. Alternatively, trajectory inference~\cite{capellera2024transportmer, kim2023ball} aims to reconstruct the full trajectory of an unobserved agent, such as the ball, using only the observed dynamics of the surrounding players. This task is particularly relevant in sports analytics, where ball tracking is often unavailable due to occlusions, fast motion, or limited camera coverage. Additionally, collecting ball tracking data in real time remains expensive and error-prone.

The SoccerNet-GSR dataset~\cite{somers2024soccernet} is composed of 200 video sequences of 30 seconds and provides millions of player positions in pitch coordinates, together with role, team, and jersey number annotations. It extends SoccerNet-Tracking~\cite{cioppa2022soccernet}, which contains bounding box annotations for players, referees, and the ball in image coordinates. However, when deriving pitch coordinates, Somers \etal~\cite{somers2024soccernet} assume that each athlete’s feet lie on the pitch plane. While this approximation is acceptable for players, it prevents accurate localization of the ball, which spends significant time in the air. As a result, ball annotations had to be removed in SoccerNet-GSR~\cite{somers2024soccernet}, making the dataset unsuitable for studying continuous ball trajectories in pitch coordinates and preventing its use for a comparison in our work.

BallRadar~\cite{kim2023ball} addresses this task by using a hierarchical framework: the model first predicts ball possessor at each frame, and then uses this intermediate result to guide ball trajectory inference. While effective in capturing high-level game structure, this sequential dependency imposes a rigid pipeline that can propagate errors from one stage to the next and limits flexibility in learning richer representations.

Capellera \etal~\cite{capellera2024transportmer} introduce TranSPORTmer, a unified transformer-based~\cite{vaswani2017attention} framework that handles ball inference alongside other tasks. It applies attention sequentially across temporal and social dimensions and achieves strong performance by modeling inter-agent dynamics holistically. However, the model operates exclusively on structured inputs, and does not incorporate any visual information. In contrast, our method explicitly incorporates visual appearance features from player-centric crops alongside positional inputs. This multi-modal formulation allows the model to make more informed inferences about ball dynamics, especially in ambiguous or low-confidence situations where structure-only methods may underperform.

\noindent\textbf{State Classification.} Classifying semantic ball state, such as passes, possessions, or uncontrolled moments, is essential for high-level soccer analytics. \cite{peral2025temporally} addresses this task through visual recognition of specific ball events. Their model first identifies the ball possessor using image-based cues and then infers high-level actions. While having a very refined temporal precision, the strict dependency on accurate ball possessor identification introduces rigid structural assumptions especially in visually challenging scenarios.

TranSPORTmer~\cite{capellera2024transportmer} incorporates state classification within a broader multi-task setup, using structured positional data as input. A CLS token~\cite{devlin2019bert} to encode frame-level information is proposed, enabling ball-state predictions in parallel with other outputs. However, the absence of visual input limits the model's ability to detect subtleties that may not be apparent from just position data, like players posture or intent. Our joint use of trajectory information and image-based player crops enhances the model’s capacity to distinguish details like body orientation or physical interactions, i.e., a better classification of fine-grained ball states.

\noindent\textbf{Masking Strategies.} Masking has become a widely adopted strategy for pre-training, particularly in NLP~\cite{devlin2019bert, wettig2022should, yang2022learning, liang2024bpdec, zheng2025exlm} and visual representation learning~\cite{dosovitskiy2020image, bao2021beit, he2022masked, tong2022videomae, bachmann2022multimae}. These approaches demonstrate that forcing models to reconstruct missing or occluded inputs encourages generalization, robustness to noise, and a deeper understanding of contextual dependencies.

MultiMAE~\cite{bachmann2022multimae} extends this idea by masking patches across multiple modalities from the visual domain (RGB, depth, and semantic segmentation) and training the model to reconstruct them from the remaining views. This design promotes cross-modal predictive coding and transferable feature representations. While the authors use random per-modality masking and report good results, they acknowledge that such a naive strategy may not always be optimal for every application or modality.

In our setting, we empirically find that sequence-level masking of the visual modality alone leads to better performance across tasks. Random frame-wise masking proves ineffective due to the high redundancy across adjacent frames and its poor alignment with real-world fails, like continuous occlusions or missing image for players out of view. Our findings support the idea that task-specific masking strategies, particularly those aligned with realistic failure scenarios, can significantly improve robustness and representation quality in multi-modal learning.

\section{Methodology}
\label{sec:method}

In this section we introduce our approach to jointly infer ball trajectory, classify the ball state, and the possessor of the ball in every frame, all of them, exploiting as input a set of heterogeneous modalities: 1) visual crops around players from the video footage, 2) the trajectory of the players in pitch coordinates and, 3) the semantics of each agent indicating which type of player they are. To start with, we show the architecture we use (see Fig.~\ref{fig:overview}) that is divided into several blocks. Particularly, our model consists of a visual input feature extraction pipeline, the processing of the remaining inputs and their fusion, some attention-based temporal and social modules, and the prediction heads. Next, we describe each of them and follow with the explanation of our pre-training masking strategy called CropDrop.

\subsection{Crops Branch}
\label{subsec:crops}

To handle visual information, we crop localized regions of the footage frame around each player~\cite{sanford2020group, sorano2020automatic, honda2022pass}. To this end, we extract bounding boxes of every player along time using MixSort~\cite{cui2023sportsmot} and enlarge them by adding a fixed margin to preserve contextual cues. These cropped regions are then passed through a convolutional neural network that acts as a feature extractor. Specifically, we employ a ResNet~\cite{he2016deep} backbone we train on an auxiliary action recognition task. This auxiliary task consists of classifying whether a player in the crop is performing a ball-related action (e.g., passing, receiving, dribbling) or not, which provides a strong inductive bias toward capturing ball-related visual cues. We chose this lightweight architecture for efficiency and reproducibility. The resulting encoder produces compact visual descriptors for each crop, which are then used as the appearance features for each player in every frame. These crop-based embeddings serve as one of the input modalities to the model.

\subsection{Other Modes and Fusion}
\label{subsec:fusion}
In addition to the visual embeddings, our model processes two structured inputs per player: their 2D position trajectories and their player type, the latter meaning whether they belong to the offensive or defensive team. The trajectories are represented as sequences of $(x,y)$ coordinates over time, and the player types are encoded as one-hot vectors. To enable joint processing of all modalities, we first project each input into a shared latent space. We then concatenate the projected embeddings along the feature dimension for each player and frame, and pass them through a small stack of fully connected layers with non-linearities to refine the fusion and prepare the data for the attention downstream.

To support ball inference and ball state classification, we append two learned CLS tokens~\cite{devlin2019bert} to the set of player representations at each timestep. This results in a tensor of shape $T \times (N+2) \times d$, where $T$ is the number of frames, $N$ the number of players, and $d$ the embedding dimension. The CLS tokens aggregate global information across both time and agents, and are supervised as part of the ball inference and ball state classification heads described later.

\subsection{Temporal and Social Attention Modules}
\label{subsec:transformers}
The fused per-player embeddings, with the appended CLS tokens, are processed by a two-stage transformer-based encoder designed to capture both temporal dynamics and social interactions~\cite{capellera2024transportmer}. This separation encourages initial rough aggregation followed by more precise contextualization, improving the ability to balance global understanding with local precision. The first encoder, referred to as coarse encoder, enables the model to capture the big picture of the scene, like getting an early sense of team formations and identifying key players. The stable representation provided by the coarse encoder is passed to the second encoder, referred as fine encoder, which performs a second round of attention-based infering to model finer-grained temporal cues and subtle inter-agent interactions.
Each encoder operates on entire $T \times (N+2) \times d$ sequence and is composed of two Temporal Set Attention Blocks (SAB$_\text{T}$) followed by a Social Set Attention Block (SAB$_\text{S}$)~\cite{lee2019set}. The encoders are preceded by the addition of temporal positional encoding~\cite{vaswani2017attention} to retain chronological order information.

The SAB$_\text{T}$ operates independently on each agent across the temporal axis. This allows the model to aggregate information from past and future frames for each entity, enabling temporal context for smoother and more informed representation of player dynamics. The SAB$_\text{S}$ then attends across all agents within each frame, modeling inter-player interactions and spatial dependencies. This includes attention from and to the CLS tokens, allowing them to integrate player-level features into a scene-level representation.

\subsection{Prediction Heads and Losses}
\label{subsec:losses}
Our model is jointly trained on three tasks: ball position inference, ball state classification and ball possessor identification. Each task is supervised with a dedicated head applied to the relevant output tokens of the transformer stack.

\noindent \textbf{Ball Position Inference.}
A key challenge of this task is that the model must infer the ball's trajectory without direct access to any past or future ball observations. The input consists solely of player-centric information (trajectories, semantic types, and visual crops) yet the model is able to infer plausible ball motion patterns based on social dynamics.
The output embedding corresponding to the ball CLS token has shape $T \times 1 \times d$, and is passed through a stack of fully connected layers with non-linear activations to produce a $T \times 2$ sequence representing the predicted $(x,y)$ ball coordinates for each frame. 

We supervise this inference using an Average Displacement Error (ADE), the mean $l_2$ distance between the predicted and ground truth ball positions across all timesteps:
\begin{equation}
  \mathcal{L}_\text{ball} = \frac{1}{T} \sum_{t=1}^T \left\| \hat{b}_t - b_t \right\|_2 ,
  \label{eq:ADE}
\end{equation}
where $\hat{b}_t$ and $b_t$ are the predicted and ground truth ball positions in 2D pitch coordinates at time $t$, respectively.

\noindent \textbf{Ball State Classification.}
To predict the state of the ball, we use the output of the state CLS token, shaped $T \times 1 \times d$. This embedding is passed through a Multi Layer Perceptron (MLP), producing $T \times S$ logits, where $S$ is the number of discrete ball state classes.
The supervision is provided by a standard cross-entropy loss between the predicted distribution and the ground truth state label at each timestep:
\begin{equation}
    \mathcal{L}_\text{state} = -\frac{1}{T} \sum_{t=1}^T \sum_{s=1}^S c_t^\text{s} \log (\hat{c}_t^\text{s}) ,
    \label{eq:state}
\end{equation}
where $\hat{c}_t^\text{s}$ denotes the predicted probability of the ball being in state $s$ at time step $t$, and $c_t^\text{s}$ is the ground truth.

\noindent \textbf{Ball Possessor Identification.} To find the player in possession of the ball, the model uses the set of player tokens of shape $T \times N \times d$. These embeddings are processed by a fully connected prediction head that produces logits of shape $T \times N$, representing the probability distribution over all $N$ players at each timestep.
We apply a cross-entropy loss that encourages the model to assign high probability to the correct ball possessor:
\begin{equation}
    \mathcal{L}_\text{poss} = -\frac{1}{T} \sum_{t=1}^T \sum_{n=1}^N c_t^\text{n} \log (\hat{c}_t^\text{n}) ,
    \label{eq:poss}
\end{equation}
where $c_t^\text{n}$ is the ground truth probability of player $n$ being the possessor of the ball at time step $t$, and $\hat{c}_t^\text{n}$ represents the predicted probability.

\noindent \textbf{Total Loss.}
The total training objective is the weighted sum of the three individual losses:
\begin{equation}
    \mathcal{L}_\text{total} = \lambda_1 \mathcal{L}_\text{ball} + \lambda_2 \mathcal{L}_\text{state} + \lambda_3 \mathcal{L}_\text{poss} ,
    \label{eq:loss}
\end{equation}
where $\lambda_1$, $\lambda_2$, and $\lambda_3$ are hyperparameters that control the relative importance of each task. They are empirically tuned to balance convergence and task-specific performance.

\subsection{CropDrop}
\label{subsec:cropdrop}
To improve the model’s ability to capture subtle visual cues we adopt CropDrop, a pre-training strategy in which the model is trained on the same multi-task objectives, but with partial masking applied only to the visual modality. The reason behind applying CropDrop just to the visual modality lies on the fact that features extracted from player image crops are often dominated by whether or not the ball is visible in the crop. As a result, most players share similar visual embeddings, except for the few who are near the ball or in possession of it. This imbalance can lead the model to over-rely on visual cues and develop a shortcut: consistently assigning possession to the player whose image features most deviate from the others. By selectively masking the visual features of random players, we prevent the model from depending exclusively on this shortcut. Instead, it is encouraged to leverage complementary information from other modalities. In this way, masking sparse but highly informative signals fosters the discovery of latent cross-modal patterns and improves generalization.

Concretely, before the modality fusion stage, we randomly select a percentage of players in each input sequence and mask their visual features across all frames. CropDrop is applied only to the crops branch modality, while preserving the input dimensionality. To do so, we replace the masked visual embeddings with fixed placeholder values that lie outside the valid range of the visual features, effectively signaling the absence of information without affecting the model architecture.

\section{Experiments}
\label{sec:experiments}
We now present both quantitative and qualitative results for the three tasks addressed by our model: ball trajectory inference, ball state classification and ball possessor identification. First of all, we describe the dataset and evaluation protocol, followed by a series of ablation studies assessing the contribution of each input modality and the impact of CropDrop, our masked visual pre-training strategy. Finally, we compare our method against two state-of-the-art methods that do not use visual information across all tasks.

Evaluation for ball inference is based on two standard metrics: the Average Displacement Error (ADE) defined in Eq.~\eqref{eq:ADE} and the Maximum Error (MaxErr) as: 
\begin{equation}
  \textrm{MaxErr} = \underset{t\in \{1,\ldots,T\}}{\textrm{max}} \left(\left\| \hat{b}_t - b_t \right\|_2 \right),
  \label{eq:maxerr}
\end{equation}
where $\hat{b}_t$ is the predicted ball position and $b_t$ is the ground truth at time step $t$. Both metrics are represented in meters.

For classification tasks (ball state and ball possessor identification) we report frame-level accuracy as a percentage, comparing the predicted class to the ground truth label at each time step. In our experiments, we set $\lambda_1=1$, $\lambda_2=3$, and $\lambda_3=3$ in Eq.~\eqref{eq:loss}, and keep a constant masking ratio for all pre-trainings of $0.5$. 

\begin{table*}
  \centering
  \resizebox{17.4cm}{!} {
  \begin{tabular}{@{}lcccccc@{}}
    \toprule
    Method & Masking & Masked modes & ADE$\downarrow$ & MaxErr$\downarrow$ & States Acc (\%)$\uparrow$ & Possessor Acc (\%)$\uparrow$ \\
    \midrule
    \midrule
    Ballradar~\cite{kim2023ball} & - & - & 1.8612 & 7.0211 & - & 78.12 \\
    TranSPORTmer~\cite{capellera2024transportmer} & - & - & 1.7179 & 6.5634 & 79.77 & 83.79 \\
    \midrule
    \multirow{7}{*}{Ours}
    & None & None & 1.4544 & 5.3678 & 84.38 & 87.67 \\
    \cline{2-7}
    & & Both & 1.5602 & 5.2958 & 84.98 & 88.71 \\
    &Random frames & Traj & 1.4829 & 5.4726 & 84.30 & 87.96 \\
    & & Crops & \underline{1.2859} & \underline{5.0142} & \underline{85.35} & \underline{88.95} \\
    \cline{2-7}
    & & Both & 1.4933 & 5.1550 & 85.29 & \underline{88.95} \\
    &All frames & Traj & 1.9008 & 5.7830 & 84.06 & 87.60 \\
    & & Crops & \textbf{1.1735} & \textbf{4.7744} & \textbf{85.85} & \textbf{89.56} \\
    \midrule
    \midrule
    Relative ratio to  & None & None & 0.8466 & 0.8178 & 1.0578 & 1.0463\\
    TranSPORTmer~\cite{capellera2024transportmer} \textsuperscript{$\dagger$}& {All frames} & Crops & \textbf{0.6831} & \textbf{0.7274} & \textbf{1.0762} & \textbf{1.0688} \\
    \bottomrule
  \end{tabular}}
  \vspace{-0.2cm}
  \caption{\textbf{Quantitative comparison and ablation study in terms of masking.} The table shows the effect of several types of masking when is applied to player crops, player trajectories, or a combination of both. The table reports some metrics for the following baselines: Ballradar~\cite{kim2023ball} and TranSPORTmer~\cite{capellera2024transportmer}; and ours. \textsuperscript{$\dagger$} Relative error or accuracy ratio is always computed between our model and TranSPORTmer~\cite{capellera2024transportmer}. Values greater than 1 indicate relative gains in accuracy, while values smaller than 1 indicate relative reductions in error.}
  \label{tab:masking}
\end{table*}

\begin{table}[t!]
  \centering
  \begin{tabular}{@{}ccc|cccc@{}}
    \toprule
    Traj & Types & Crops & ADE$\downarrow$ & ME$\downarrow$ & SA $\uparrow$ & PA $\uparrow$ \\
    \midrule
    \midrule
    \XSolidBrush & \XSolidBrush & \Checkmark & 23.951 & 35.769 & 77.01 & 81.96 \\
    \XSolidBrush & \Checkmark & \Checkmark & 19.933 & 31.755 & 77.54 & 82.00 \\
    \Checkmark & \XSolidBrush & \XSolidBrush & 1.9749 & 7.1197 & 77.79 & 82.12 \\
    \Checkmark & \Checkmark & \XSolidBrush & 1.7814 & 6.5516 & 79.64 & 83.66 \\
    \Checkmark & \XSolidBrush & \Checkmark & 1.6751 & 5.8967 & 82.89 & 86.38 \\
    \Checkmark & \Checkmark & \Checkmark & \textbf{1.4544} & \textbf{5.3678} & \textbf{84.38} & \textbf{87.67} \\
    \bottomrule
  \end{tabular}
  \vspace{-0.2cm}
  \caption{\textbf{Quantitative ablation on multi-modality.} The table reports ADE and ME, in meters, for ball trajectory and both State (SA) and Possessor Accuracy (PA). It is provided the effect of every modality, being the last row the full model we propose.}
  \label{tab:modes}
\end{table}

\subsection{Dataset}
\label{subsec:dataset}
Our experiments are conducted on a proprietary soccer dataset comprising 283 full professional matches from a top European league during season 2022-2023. 
Each frame is accompanied by a tactical camera (Camera 1) image. This camera captures a wide-angle view that follows the flow of the game, typically ensuring visibility of most outfield players, although the goalkeeper at the opposite side of the field is usually out of view. Additionally, for each frame, we precompute axis-aligned bounding boxes for all visible players, used to generate the cropped image inputs for the visual modality. Every player is also associated with a team identity, which is used as the player type input to the model.

We are provided with ground truth 2D player trajectories in pitch coordinates and we also have access to the ball possessor at every timestep. Using possessor annotations, we interpolate the 2D trajectory of the ball. Each frame is annotated with one of four possible ball state labels: \textit{pass}, \textit{possession}, \textit{uncontrolled}, and \textit{out of play}. We exclude sequences with frames labeled as \textit{out of play}, as they are not informative for modeling ball dynamics. During these intervals, the flow of the match is paused: players may be walking, repositioning, or even carrying the ball with their hands, and the coordinated team behavior such as formations, pressing structures, or ball-oriented reactions completely breaks down. Since our focus is on learning from coordinated play and real-time interactions, it is both intuitive and necessary to discard these sequences.

From the matches, we extract sequences of 60 frames sampled at 6.25 Hz, corresponding to 9.6 seconds of gameplay per sequence. The dataset is split into training, validation, and test sets containing $43,397$, $3,968$, and $3,347$ sequences, respectively. Matches are disjoint across splits to ensure generalization to unseen games. As per the instances for each ball state in the dataset, the number is $82,961$ pass frames, $97,841$ possession ones and $20,018$ uncontrolled.

\begin{figure*}[t!]
  \centering
   \includegraphics[width=1\linewidth]{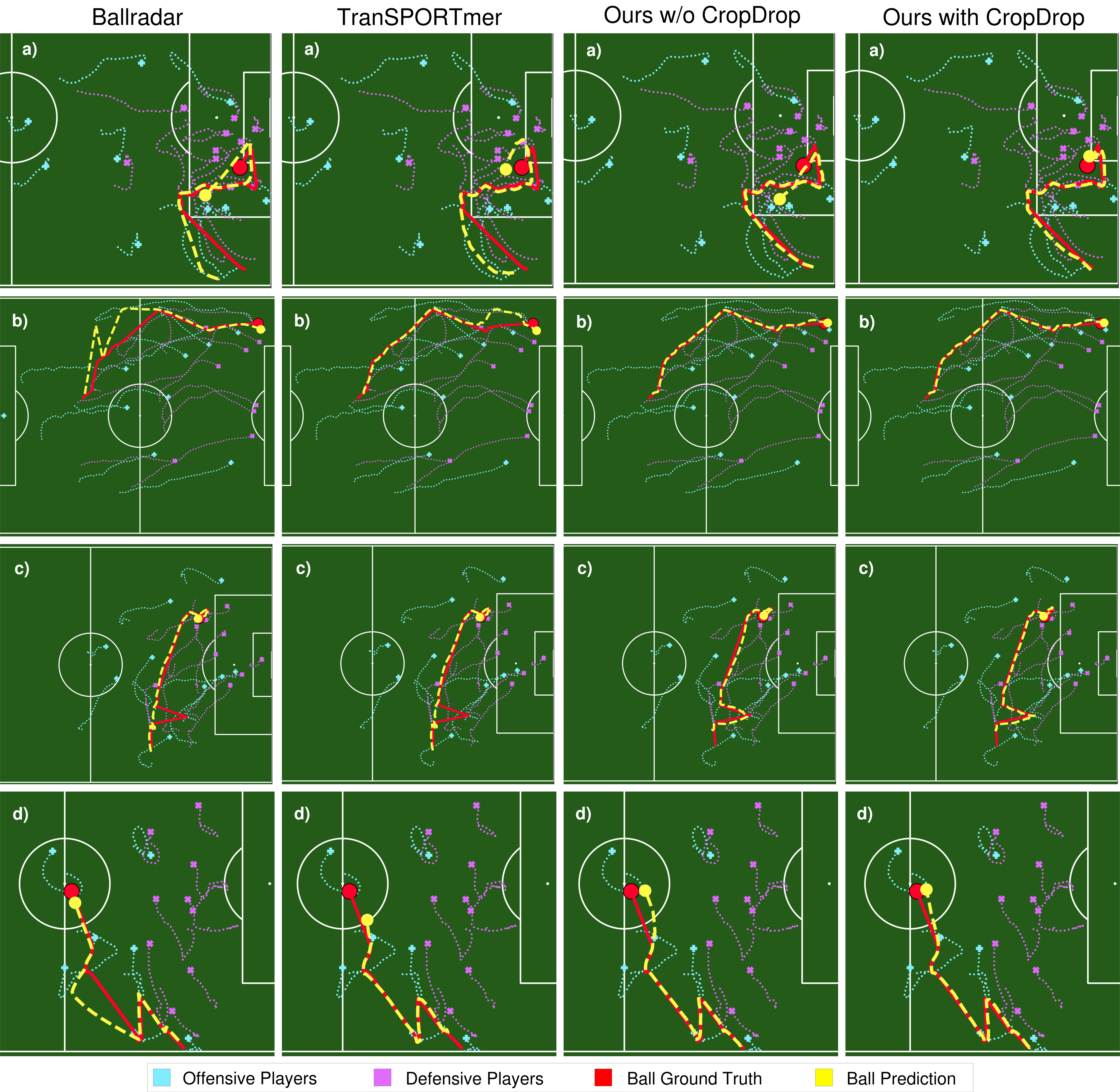}
   \caption{\textbf{Qualitative evaluation.} The figure shows 4 sequences where both offensive and defensive players are displayed with the ball prediction and the corresponding ground truth motion. Markers represent the ending point of trajectories. Sequence a) depicts our model superiority in ball inference. As some passes are incorrectly hallucinated by competing approaches, ball trajectory is wrong for them, as evidenced in b). In c) competing approaches miss the one-two pass at the start of the sequence while our model correctly captures it. d) manifests that when a sequence begins or ends with the ball not in possession, our model struggles due to limited temporal context.
  }
   \label{fig:c2}
\end{figure*}

\subsection{Analyzing modalities}
\label{subsec:multimodal}
We begin our evaluation by analyzing the contribution of each input modality. \Cref{tab:modes} reports results across all three downstream tasks: ball trajectory inference, ball state classification, and possessor identification.
Using only trajectories provides a strong baseline with an ADE of 1.97 and MaxError (ME) of 7.12 in ball inference, along with reasonable classification accuracy for both state (77.79\%) and possessor (82.12\%) is obtained.
Trajectory information is thus critical, since the other modalities are insufficient to reconstruct the dynamic ball motion. Despite that, visual crops still contribute meaningfully to the classification tasks, also reaching reasonable accuracies on their own, confirming their utility in capturing subtle visual indicators such as ball control. Adding player-type information further improves performance across all tasks. The additional semantic context helps the model better differentiate between teams compared to the only trajectories baseline, contributing in an ADE reduction of more than 0.5 meters and almost a +7\% and +6\% in state and possessor accuracy, respectively.

Combining trajectories with crops yields substantial improvement across the board, showing that visual features provide valuable complementary information for disambiguating complex scenes or subtle interactions between players. Finally, the full model using all three modalities achieves the best performance, confirming the synergistic effect of combining visual and structure modalities in a shared representation space.
This ablation study validates that multi-modal fusion significantly enhances both physical inference (ball trajectory) and semantic understanding (state and possession), with each input stream contributing distinct and complementary information to the model.

\subsection{Pre-Train with CropDrop}
\label{subsec:masking_res}

We now evaluate the impact of CropDrop, our visual masking strategy during pre-training, which was introduced above. Our goal is to show how selectively masking the visual modality during pre-training improves model robustness and overall performance across all tasks. \Cref{tab:masking} presents results for various masking configurations. First, we compare two temporal masking strategies: masking random frames versus masking full sequences for randomly selected players. We observe that per-player full-sequence masking consistently outperforms sparse frame masking. This aligns with the intuition that randomly masking individual frames is ineffective due to the high redundancy across adjacent frames at a dense frame rate. Moreover, full-sequence masking better reflects real-world missing data scenarios, such as players being occluded or moving out of the camera’s field of view for several seconds.

Next, we analyze the impact of which modalities are masked. We refrain from masking the player type input given its simplicity and low information content, considering that masking this feature will unlikely offer meaningful regularization benefits and may instead increase ambiguity during training. Furthermore, this input is always available in real-world scenarios, as player team affiliation is consistently known across all frames and does not depend on visual coverage, sensor data, or tracking systems that may fail or degrade. When masking only trajectories or both trajectories and visual crops performance degrades. We hypothesize that trajectory masking deprives the model of essential structural information, making the learning unstable or ineffective.
In contrast, masking only the visual modality proves beneficial. This forces the model to rely more on the structured inputs during pre-training, encouraging it to generalize under partial input conditions. When fine-tuned with access to full visual data, the model can then exploit the visual information more effectively, having learned to compensate for its absence during pre-training. This leads to the best performance across all metrics: 1.17 ADE, 4.77 MaxErr, 85.85\% state classification accuracy, and 89.56\% possessor identification accuracy. These findings support our central claim: CropDrop as a task-aligned, modality-specific masking in the visual branch acts as an effective regularizer, improving both generalization and downstream task performance in multi-modal learning.

\subsection{Evaluation and Comparison}
\label{subsec:sota}

We compare our method against two recent state-of-the-art approaches for ball trajectory inference and semantic understanding in multi-agent sports settings: BallRadar~\cite{kim2023ball} and TranSPORTmer~\cite{capellera2024transportmer}. It is worth noting that both methods do not use player crops as input, as we do in this paper. To better illustrate improvements we also report relative error and accuracy values in~\Cref{tab:masking}, computed as ratios with respect to TranSPORTmer~\cite{capellera2024transportmer}. This allows us to express gains directly, e.g., 1.0762 in state accuracy means a +7.62\% improvement over TranSPORTmer. Our approach consistently outperforms both baselines, achieving the lowest ADE and MaxErr of 1.17 and 4.77 meters in ball trajectory inference, i.e, reducing ADE and MaxErr by 31.69\% and 27.26\%, respectively, with respect to TranSPORTmer~\cite{capellera2024transportmer}, the most accurate approach in the literature. Some qualitative results can be seen in Fig.~\ref{fig:c2}a), observing how the use of CropDrop is a key factor in obtaining more accurate predictions.

In addition to spatial accuracy, our model also demonstrates superior semantic understanding. While Ballradar~\cite{kim2023ball} and TranSPORTmer~\cite{capellera2024transportmer} achieve competitive results through structured attention, our model improves performance, reaching 85.85\% accuracy for state classification and 89.56\% for possessor identification, i.e., an increase of 7.62\% and 6.88\% with respect to TranSPORTmer~\cite{capellera2024transportmer}. As can be seen in Fig.~\ref{fig:c2}b), Ballradar~\cite{kim2023ball} hallucinates four passes at the beginning of the sequence, and TranSPORTmer~\cite{capellera2024transportmer} two at the end. During the frames where those passes are present in their predictions, both approaches guess a wrong possessor of the ball, and TranSPORTmer~\cite{capellera2024transportmer} predicts an incorrect ball state assigning a pass when there is a player in possession driving the ball. In contrast, our solution reduces this type of errors. The opposite error is shown in Fig.~\ref{fig:c2}c). In this case, the competing methods miss a one-two pass, or wall pass, while our model is capable of following the ball during this fast pass performed by two players in combination to get past a defender.

Although our method shows strong overall performance, some limitations remain. As illustrated in Fig.~\ref{fig:c2}d), when a sequence ends with the ball traveling between two players, the model lacks sufficient temporal context to confidently determine its final location. This often leads to errors in the last part of the predicted trajectory, despite accurate tracking earlier in the clip. These situations highlight the importance of longer temporal context and suggest future work extending the model to operate over extended clips.

\section{Conclusion}
\label{sec:conclusions}

We have introduced a multi-modal transformer-based architecture for soccer scene understanding, capable of jointly inferring ball trajectories, classifying game states, and determining the ball possessor at each frame. Our approach fuses structural inputs (trajectories and player types) with visual information (player-centric image crops) interpreting complex in-game interactions.
To improve generalization and robustness we introduce CropDrop, a novel visual masking pre-training strategy that hides player crops during training to simulate occlusion and prevent over-reliance on image features. This forces the model to develop stronger cross-modal judging capabilities, enabling it to better handle ambiguous or incomplete inputs at inference time. With experiments through an extensive dataset of professional matches, we demonstrated that each input modality contributes meaningfully to performance, and that our masking strategy provides consistent gains across tasks. The upgrade with visual information is proven against state-of-the-art baselines. Our method achieves superior accuracy in both ball state classification and possessor identification, and a lower error in ball trajectory inference.
Further work could exploit the addition of extra inputs or even try to leverage more tasks jointly to the already existing ones.

\noindent \textbf{Acknowledgment.} This work has been partially supported by the project GRAVATAR PID2023-151184OB-I00 funded by MCIU/AEI/10.13039/501100011033 and by ERDF, UE and by the Government of Catalonia under 2020 DI 00106.

{
    \small
    \bibliographystyle{ieeenat_fullname}
    \bibliography{main}

@String(CVPR= {IEEE Conf. Comput. Vis. Pattern Recog.})

@String(ICCV= {Int. Conf. Comput. Vis.})

@String(ECCV= {Eur. Conf. Comput. Vis.})

@String(TVCG  = {IEEE Trans. Vis. Comput. Graph.})

@String(ICIP = {IEEE Int. Conf. Image Process.})

@String(ACCV  = {ACCV})

@String(ICLR = {Int. Conf. Learn. Represent.})

@String(CVPRW= {IEEE Conf. Comput. Vis. Pattern Recog. Worksh.})

@String(CVPR  = {CVPR})

@String(ICCV  = {ICCV})

@String(ECCV  = {ECCV})

@String(TVCG  = {IEEE TVCG})

@String(ICIP  = {ICIP})

@String(ICLR  = {ICLR})

@String(CVPRW= {CVPRW})

@article{amirli2022prediction,
  title={Prediction of the ball location on the {2D} plane in football using optical tracking data},
  author={Amirli, Anar and Alemdar, Hande},
  journal={APJESS},
  volume={10},
  number={1},
  pages={1--8},
  year={2022}
}

@inproceedings{kim2023ball,
  title={Ball trajectory inference from multi-agent sports contexts using set transformer and hierarchical bi-lstm},
  author={Kim, Hyunsung and Choi, Han-Jun and Kim, Chang Jo and Yoon, Jinsung and Ko, Sang-Ki},
  booktitle={SIGKDD},
  pages={4296--4307},
  year={2023}
}

@inproceedings{capellera2024transportmer,
  title={TranSPORTmer: A Holistic Approach to Trajectory Understanding in Multi-Agent Sports},
  author={Capellera, Guillem and Ferraz, Luis and Rubio, Antonio and Agudo, Antonio and Moreno-Noguer, Francesc},
  booktitle={ACCV},
  pages={1652--1670},
  year={2024}
}

@article{peral2025temporally,
  title={Temporally accurate events detection through ball possessor recognition in soccer},
  author={Peral, Marc and Capellera, Guillem and Rubio, Antonio and Ferraz, Luis and Moreno-Noguer, Francesc and Agudo, Antonio},
  journal={VISAPP},
  volume={221},
  pages={231},
  year={2025}
}

@inproceedings{soccer3D,
  title={SoccerNet-v{3D}: Leveraging Sports Broadcast Replays for {3D} Scene Understanding},
  author={Gutiérrez-Pérez, Marc and Agudo, Antonio},
  booktitle={CVPRW},
  year={2025}
}

@inproceedings{perezyusWACV22,
  title={Matching and Recovering {3D} People from Multiple Views},
  author={Perez-Yus, Alejandro and Agudo, Antonio},
  booktitle={WACV},
  pages={1184--1193},
  year={2022}
}

@inproceedings{sorano2020automatic,
  title={Automatic pass annotation from soccer video streams based on object detection and lstm},
  author={Sorano, Danilo and Carrara, Fabio and Cintia, Paolo and Falchi, Fabrizio and Pappalardo, Luca},
  booktitle={ECML PKDD},
  pages={475--490},
  year={2020}
}

@article{gutierrez4998149pnlcalib,
  title={Pnlcalib: Sports Field Registration Via Points and Lines Optimization},
  year={2025}, 
  author={Guti{\'e}rrez-P{\'e}rez, Marc and Agudo, Antonio},
  journal={Available at SSRN 4998149}
}

@article{vidal2022automatic,
  title={Automatic event detection in football using tracking data},
  author={Vidal-Codina, Ferran and Evans, Nicolas and El Fakir, Bahaeddine and Billingham, Johsan},
  journal={Sports Engineering},
  volume={25},
  number={1},
  pages={18},
  year={2022},
  publisher={Springer}
}

@article{link2017individual,
  title={Individual ball possession in soccer},
  author={Link, Daniel and Hoernig, Martin},
  journal={PloS one},
  volume={12},
  number={7},
  pages={e0179953},
  year={2017},
  publisher={Public Library of Science San Francisco, CA USA}
}

@inproceedings{devlin2019bert,
  title={Bert: Pre-training of deep bidirectional transformers for language understanding},
  author={Devlin, Jacob and Chang, Ming-Wei and Lee, Kenton and Toutanova, Kristina},
  booktitle={NAACL-HLT},
  pages={4171--4186},
  year={2019}
}

@article{vaswani2017attention,
  title={Attention is all you need},
  author={Vaswani, Ashish and Shazeer, Noam and Parmar, Niki and Uszkoreit, Jakob and Jones, Llion and Gomez, Aidan N and Kaiser, {\L}ukasz and Polosukhin, Illia},
  journal={NeurIPS},
  volume={30},
  year={2017}
}

@article{majeed2025real,
  title={Real-time analysis of soccer ball--player interactions using graph convolutional networks for enhanced game insights},
  author={Majeed, Fahad and Nazir, Maria and Swart, Kamilla and Agus, Marco and Schneider, Jens},
  journal={Scientific Reports},
  volume={15},
  number={1},
  pages={21859},
  year={2025},
  publisher={Nature Publishing Group UK London}
}

@article{komorowski2019deepball,
  title={Deepball: Deep neural-network ball detector},
  author={Komorowski, Jacek and Kurzejamski, Grzegorz and Sarwas, Grzegorz},
  journal={VISAPP},
  year={2019}
}

@article{yamamoto2024theory,
  title={Theory and data analysis of player and team ball possession time in football},
  author={Yamamoto, Ken and Uezu, Seiya and Kagawa, Keiichiro and Yamazaki, Yoshihiro and Narizuka, Takuma},
  journal={Physical Review E},
  volume={109},
  number={1},
  pages={014305},
  year={2024},
  publisher={APS}
}

@article{ochin2025game,
  title={Game State and Spatio-temporal Action Detection in Soccer using Graph Neural Networks and {3D} Convolutional Networks},
  author={Ochin, Jeremie and Devineau, Guillaume and Stanciulescu, Bogdan and Manitsaris, Sotiris},
  journal={ICPRAM},
  year={2025}
}

@inproceedings{capellera2024footbots,
  title={Footbots: A transformer-based architecture for motion prediction in soccer},
  author={Capellera, Guillem and Ferraz, Luis and Rubio, Antonio and Agudo, Antonio and Moreno-Noguer, Francesc},
  booktitle={ICIP},
  pages={2313--2319},
  year={2024}
}

@inproceedings{sanford2020group,
  title={Group activity detection from trajectory and video data in soccer},
  author={Sanford, Ryan and Gorji, Siavash and Hafemann, Luiz G and Pourbabaee, Bahareh and Javan, Mehrsan},
  booktitle={CVPRW},
  pages={898--899},
  year={2020}
}

@inproceedings{capellera2025unified,
  title={Unified uncertainty-aware diffusion for multi-agent trajectory modeling},
  author={Capellera, Guillem and Rubio, Antonio and Ferraz, Luis and Agudo, Antonio},
  booktitle={CVPR},
  pages={22476--22486},
  year={2025}
}

@inproceedings{bachmann2022multimae,
  title={Multimae: Multi-modal multi-task masked autoencoders},
  author={Bachmann, Roman and Mizrahi, David and Atanov, Andrei and Zamir, Amir},
  booktitle={ECCV},
  pages={348--367},
  year={2022}
}

@article{zheng2025exlm,
  title={ExLM: Rethinking the Impact of [MASK] Tokens in Masked Language Models},
  author={Zheng, Kangjie and Yang, Junwei and Liang, Siyue and Feng, Bin and Liu, Zequn and Ju, Wei and Xiao, Zhiping and Zhang, Ming},
  journal={ICML},
  year={2025}
}

@article{wettig2022should,
  title={Should you mask 15\% in masked language modeling?},
  author={Wettig, Alexander and Gao, Tianyu and Zhong, Zexuan and Chen, Danqi},
  journal={EACL},
  pages={2985--3000},
  year={2022}
}

@article{yang2022learning,
  title={Learning better masking for better language model pre-training},
  author={Yang, Dongjie and Zhang, Zhuosheng and Zhao, Hai},
  journal={ACL},
  year={2023},
  pages={7255--7267}
}

@article{liang2024bpdec,
  title={BPDec: unveiling the potential of masked language modeling decoder in BERT pretraining},
  author={Liang, Wen and Liang, Youzhi},
  journal={arXiv preprint arXiv:2401.15861},
  year={2024}
}

@inproceedings{he2022masked,
  title={Masked autoencoders are scalable vision learners},
  author={He, Kaiming and Chen, Xinlei and Xie, Saining and Li, Yanghao and Doll{\'a}r, Piotr and Girshick, Ross},
  booktitle={CVPR},
  pages={16000--16009},
  year={2022}
}

@article{tong2022videomae,
  title={Videomae: Masked autoencoders are data-efficient learners for self-supervised video pre-training},
  author={Tong, Zhan and Song, Yibing and Wang, Jue and Wang, Limin},
  journal={NeurIPS},
  volume={35},
  pages={10078--10093},
  year={2022}
}

@article{dosovitskiy2020image,
  title={An image is worth 16x16 words: Transformers for image recognition at scale},
  author={Dosovitskiy, Alexey and Beyer, Lucas and Kolesnikov, Alexander and Weissenborn, Dirk and Zhai, Xiaohua and Unterthiner, Thomas and Dehghani, Mostafa and Minderer, Matthias and Heigold, Georg and Gelly, Sylvain and others},
  journal={ICLR},
  year={2021}
}

@article{bao2021beit,
  title={Beit: Bert pre-training of image transformers},
  author={Bao, Hangbo and Dong, Li and Piao, Songhao and Wei, Furu},
  journal={ICLR},
  year={2022}
}

@inproceedings{honda2022pass,
  title={Pass receiver prediction in soccer using video and players' trajectories},
  author={Honda, Yutaro and Kawakami, Rei and Yoshihashi, Ryota and Kato, Kenta and Naemura, Takeshi},
  booktitle={CVPR},
  pages={3503--3512},
  year={2022}
}

@inproceedings{lee2019set,
  title={Set transformer: A framework for attention-based permutation-invariant neural networks},
  author={Lee, Juho and Lee, Yoonho and Kim, Jungtaek and Kosiorek, Adam and Choi, Seungjin and Teh, Yee Whye},
  booktitle={ICML},
  pages={3744--3753},
  year={2019}
}

@inproceedings{felsen2018will,
  title={Where will they go? predicting fine-grained adversarial multi-agent motion using conditional variational autoencoders},
  author={Felsen, Panna and Lucey, Patrick and Ganguly, Sujoy},
  booktitle={ECCV},
  pages={732--747},
  year={2018}
}

@article{zhan2018generating,
  title={Generating multi-agent trajectories using programmatic weak supervision},
  author={Zhan, Eric and Zheng, Stephan and Yue, Yisong and Sha, Long and Lucey, Patrick},
  journal={ICLR},
  year={2019}
}

@article{zheng2016generating,
  title={Generating long-term trajectories using deep hierarchical networks},
  author={Zheng, Stephan and Yue, Yisong and Hobbs, Jennifer},
  journal={NeurIPS},
  volume={29},
  year={2016}
}

@article{choi2024trajectory,
  title={Trajectory Imputation in Multi-Agent Sports with Derivative-Accumulating Self-Ensemble},
  author={Choi, Han-Jun and Kim, Hyunsung and Lee, Minho and Jeong, Minchul and Kim, Chang-Jo and Yoon, Jinsung and Ko, Sang-Ki},
  journal={arXiv preprint arXiv:2408.10878},
  year={2024}
}

@article{everett2023inferring,
  title={Inferring player location in sports matches: Multi-agent spatial imputation from limited observations},
  author={Everett, Gregory and Beal, Ryan J and Matthews, Tim and Early, Joseph and Norman, Timothy J and Ramchurn, Sarvapali D},
  journal={AAMAS},
  year={2023}
}

@article{xu2024sports,
  title={Sports-traj: A unified trajectory generation model for multi-agent movement in sports},
  author={Xu, Yi and Fu, Yun},
  journal={ICLR},
  year={2025}
}

@inproceedings{xu2023uncovering,
  title={Uncovering the missing pattern: Unified framework towards trajectory imputation and prediction},
  author={Xu, Yi and Bazarjani, Armin and Chi, Hyung-gun and Choi, Chiho and Fu, Yun},
  booktitle={CVPR},
  pages={9632--9643},
  year={2023}
}

@article{omidshafiei2021time,
  title={Time-series imputation of temporally-occluded multiagent trajectories},
  author={Omidshafiei, Shayegan and Hennes, Daniel and Garnelo, Marta and Tarassov, Eugene and Wang, Zhe and Elie, Romuald and Connor, Jerome T and Muller, Paul and Graham, Ian and Spearman, William and others},
  journal={ICML},
  year={2022}
}

@article{li2025spatiotemporal,
  title={A Spatiotemporal Graph Transformer Network for real-time ball trajectory monitoring and prediction in dynamic sports environments},
  author={Li, Zujian and Yu, Dan},
  journal={Alexandria Engineering Journal},
  volume={119},
  pages={246--258},
  year={2025},
  publisher={Elsevier}
}

@inproceedings{yeh2019diverse,
  title={Diverse generation for multi-agent sports games},
  author={Yeh, Raymond A and Schwing, Alexander G and Huang, Jonathan and Murphy, Kevin},
  booktitle={CVPR},
  pages={4610--4619},
  year={2019}
}

@article{cao2018brits,
  title={Brits: Bidirectional recurrent imputation for time series},
  author={Cao, Wei and Wang, Dong and Li, Jian and Zhou, Hao and Li, Lei and Li, Yitan},
  journal={NeurIPS},
  volume={31},
  year={2018}
}

@article{liu2019naomi,
  title={Naomi: Non-autoregressive multiresolution sequence imputation},
  author={Liu, Yukai and Yu, Rose and Zheng, Stephan and Zhan, Eric and Yue, Yisong},
  journal={NeurIPS},
  volume={32},
  year={2019}
}

@inproceedings{qi2020imitative,
  title={Imitative non-autoregressive modeling for trajectory forecasting and imputation},
  author={Qi, Mengshi and Qin, Jie and Wu, Yu and Yang, Yi},
  booktitle={CVPR},
  pages={12736--12745},
  year={2020}
}

@article{gu2017non,
  title={Non-autoregressive neural machine translation},
  author={Gu, Jiatao and Bradbury, James and Xiong, Caiming and Li, Victor O.K. and Socher, Richard},
  journal={arXiv preprint arXiv:1711.02281},
  year={2017}
}

@article{lee2018deterministic,
  title={Deterministic non-autoregressive neural sequence modeling by iterative refinement},
  author={Lee, Jason and Mansimov, Elman and Cho, Kyunghyun},
  journal={arXiv preprint arXiv:1802.06901},
  year={2018}
}

@article{dick2022can,
  title={Who can receive the pass? A computational model for quantifying availability in soccer},
  author={Dick, Uwe and Link, Daniel and Brefeld, Ulf},
  journal={ECML PKDD},
  volume={36},
  number={3},
  pages={987--1014},
  year={2022}
}

@inproceedings{rahimian2023pass,
  title={Pass receiver and outcome prediction in soccer using temporal graph networks},
  author={Rahimian, Pegah and Kim, Hyunsung and Schmid, Marc and Toka, Laszlo},
  booktitle={MLSA},
  pages={52--63},
  year={2023}
}

@inproceedings{spearman2017physics,
  title={Physics-based modeling of pass probabilities in soccer},
  author={Spearman, William and Basye, Austin and Dick, Greg and Hotovy, Ryan and Pop, Paul},
  booktitle={Proceeding of the 11th MIT Sloan Sports Analytics Conference},
  volume={1},
  year={2017},
  organization={Boston, MA}
}

@article{xie2020passvizor,
  title={PassVizor: Toward better understanding of the dynamics of soccer passes},
  author={Xie, Xiao and Wang, Jiachen and Liang, Hongye and Deng, Dazhen and Cheng, Shoubin and Zhang, Hui and Chen, Wei and Wu, Yingcai},
  journal={TVCG},
  volume={27},
  number={2},
  pages={1322--1331},
  year={2020}
}

@inproceedings{cui2023sportsmot,
  title={Sportsmot: A large multi-object tracking dataset in multiple sports scenes},
  author={Cui, Yutao and Zeng, Chenkai and Zhao, Xiaoyu and Yang, Yichun and Wu, Gangshan and Wang, Limin},
  booktitle={ICCV},
  pages={9921--9931},
  year={2023}
}

@inproceedings{cioppa2022soccernet,
  title={Soccernet-tracking: Multiple object tracking dataset and benchmark in soccer videos},
  author={Cioppa, Anthony and Giancola, Silvio and Deliege, Adrien and Kang, Le and Zhou, Xin and Cheng, Zhiyu and Ghanem, Bernard and Van Droogenbroeck, Marc},
  booktitle={CVPRW},
  pages={3491--3502},
  year={2022}
}

@inproceedings{somers2024soccernet,
  title={SoccerNet game state reconstruction: End-to-end athlete tracking and identification on a minimap},
  author={Somers, Vladimir and Joos, Victor and Cioppa, Anthony and Giancola, Silvio and Ghasemzadeh, Seyed Abolfazl and Magera, Floriane and Standaert, Baptiste and Mansourian, Amir M and Zhou, Xin and Kasaei, Shohreh and others},
  booktitle={CVPR},
  pages={3293--3305},
  year={2024}
}

@article{wang2024tacticai,
  title={TacticAI: an AI assistant for football tactics},
  author={Wang, Zhe and Veli{\v{c}}kovi{\'c}, Petar and Hennes, Daniel and Toma{\v{s}}ev, Nenad and Prince, Laurel and Kaisers, Michael and Bachrach, Yoram and Elie, Romuald and Wenliang, Li Kevin and Piccinini, Federico and others},
  journal={Nature communications},
  volume={15},
  number={1},
  pages={1906},
  year={2024},
  publisher={Nature Publishing Group UK London}
}

@article{rao2024matchtime,
  title={Matchtime: Towards automatic soccer game commentary generation},
  author={Rao, Jiayuan and Wu, Haoning and Liu, Chang and Wang, Yanfeng and Xie, Weidi},
  journal={EMNLP},
  year={2024}
}

@inproceedings{rao2025towards,
  title={Towards universal soccer video understanding},
  author={Rao, Jiayuan and Wu, Haoning and Jiang, Hao and Zhang, Ya and Wang, Yanfeng and Xie, Weidi},
  booktitle={CVPR},
  pages={8384--8394},
  year={2025}
}

@article{rao2025multi,
  title={Multi-agent system for comprehensive soccer understanding},
  author={Rao, Jiayuan and Li, Zifeng and Wu, Haoning and Zhang, Ya and Wang, Yanfeng and Xie, Weidi},
  journal={ACM Multimedia},
  year={2025}
}

@inproceedings{cioppa2021camera,
  title={Camera calibration and player localization in soccernet-v2 and investigation of their representations for action spotting},
  author={Cioppa, Anthony and Deliege, Adrien and Magera, Floriane and Giancola, Silvio and Barnich, Olivier and Ghanem, Bernard and Van Droogenbroeck, Marc},
  booktitle={CVPRW},
  pages={4537--4546},
  year={2021}
}

@article{cioppa2022scaling,
  title={Scaling up SoccerNet with multi-view spatial localization and re-identification},
  author={Cioppa, Anthony and Deliege, Adrien and Giancola, Silvio and Ghanem, Bernard and Van Droogenbroeck, Marc},
  journal={Scientific data},
  volume={9},
  number={1},
  pages={355},
  year={2022},
  publisher={Nature Publishing Group UK London}
}

@inproceedings{giancola2023towards,
  title={Towards active learning for action spotting in association football videos},
  author={Giancola, Silvio and Cioppa, Anthony and Georgieva, Julia and Billingham, Johsan and Serner, Andreas and Peek, Kerry and Ghanem, Bernard and Van Droogenbroeck, Marc},
  booktitle={CVPRW},
  pages={5098--5108},
  year={2023}
}

@inproceedings{he2016deep,
  title={Deep residual learning for image recognition},
  author={He, Kaiming and Zhang, Xiangyu and Ren, Shaoqing and Sun, Jian},
  booktitle={CVPR},
  pages={770--778},
  year={2016}
}
}

\end{document}